\pdfoutput=1

\documentclass[11pt]{article}

\usepackage[preprint]{acl}

\usepackage{times}
\usepackage{latexsym}

\usepackage{amsmath,amssymb,amsfonts}
\usepackage{algorithmic}
\usepackage{graphicx}
\usepackage{textcomp}
\usepackage{xcolor}
\usepackage[ruled,linesnumbered]{algorithm2e}

\usepackage{amsmath}
\usepackage{graphicx}
\usepackage[subfigure]{graphfig}
\usepackage{multirow}
\usepackage{url}
\usepackage{booktabs} 

\usepackage[T1]{fontenc}

\usepackage[utf8]{inputenc}

\usepackage{microtype}

\usepackage{inconsolata}

\usepackage{graphicx}

\title{Neural-Bayesian Program Learning for Few-shot Dialogue Intent Parsing}

\author{Mengze Hong$^{1}$, Di Jiang$^{2}$, Yuanfeng Song$^{2}$,
Chen Jason Zhang$^{1}$\\
Hong Kong Polytechnic University$^{1}$ \\AI Group, WeBank Co., Ltd$^{2}$ \\
}

\begin{document}
\maketitle

\begin{abstract}
With the growing importance of customer service in contemporary business, recognizing the intents behind service dialogues has become essential for the strategic success of enterprises. However, the nature of dialogue data varies significantly across different scenarios, and implementing an intent parser for a specific domain often involves tedious feature engineering and a heavy workload of data labeling. In this paper, we propose a novel Neural-Bayesian Program Learning model named Dialogue-Intent Parser (\emph{DI-Parser}), which specializes in intent parsing under data-hungry settings and offers promising performance improvements. DI-Parser effectively utilizes data from multiple sources in a ``Learning to Learn'' manner and harnesses the ``wisdom of the crowd'' through few-shot learning capabilities on human-annotated datasets. Experimental results demonstrate that DI-Parser outperforms state-of-the-art deep learning models and offers practical advantages for industrial-scale applications.
\end{abstract}

\section{Introduction}

Intents are semantic labels that accurately describe the participants' intentions \cite{jiang2016query}. Intent parsing of customer service dialogue provides important insights into what customers need and how company employees respond to those needs (see Table~\ref{fig:example}), paving the way for strategies that enhance revenue and improve customer retention for enterprises \cite{harrigan2018customer}. Furthermore, accurate intent parsing is a crucial component of many downstream natural language processing (NLP) applications, such as topic mining \cite{li2021heterogeneous, jiang2021familia, jiang2016latent}, dialogue generation \cite{cui2019dal}, and natural language understanding \cite{jiang2015sg, yan2017building}.

\begin{table*}[t!]
\centering
\small
\begin{tabular}{p{11cm}|p{4cm}}
  \toprule
  \textbf{Utterance} & \textbf{Intent} \\
  \midrule
  Customer: I would like to find and book for a Vegan bistro in San Francisco for lunch & Restaurant reservation   \\
  Service: Hello. What day would you like you reservation?  greeting & Request date    \\
  Customer: March 23rd  & inform date \\
  Service:  Great, what time would you like to go to lunch? & Request start time \\
  Customer: around 1 pm & inform start time \\
  Service:  I'm happy to assist you! How many guests are in your party? & Request number of people \\
  Customer: 4 inform number of people \\
  Service:  I found three places you may like. Gracias Madre, Krua Thai, and North India Restaurant. Which would you like to book?  & Inform restaurant name, multiple choice \\
  Customer: North India restaurant would be perfect & Inform restaurant name \\
  Service:  Thank you so much for booking with us! Your reservation has been confirmed as requested. Please enjoy your meal! thanks. & Inform task complete \\
  \bottomrule
\end{tabular}
\caption{A snippet of customer service dialogue and associated intents}
\label{fig:example}
\end{table*}

Numerous studies have focused on dialogue intent parsing, encompassing both traditional classification-based approaches \cite{grau2004dialogue,stolcke2000dialogue} and more recent neural network-based methods \cite{ji2016latent,khanpour2016dialogue,kato2017utterance,yann2014zero}. Despite their advantages, existing intent parsing techniques face several challenges: (1) They require a predefined intent taxonomy \cite{perevalov2019question}; (2) a substantial amount of labeled data is needed \cite{firdaus2018deep}, making data-hungry deep neural network methods costly to train; and (3) the trained intent parsers are tightly coupled with the intent taxonomy, making it difficult to adapt them to new scenarios.

In this paper, we propose a Neural-Bayesian Program Learning (NBPL) model \cite{lake2015human} named Dialogue Intent Parser (\textbf{DI-Parser}) to address the aforementioned challenges. The core idea of NBPL is to model dialogue data through a hybrid approach that seamlessly integrates Bayesian networks and neural networks, enabling parameter fitting with limited training data. Specifically, we recognize that dialogue exhibits latent structures that can be decomposed into several compositional components, which are inherently transferable across different scenarios.

The proposed DI-Parser aims to learn and identify intents that generalize effectively from sparse data. It is formulated as a Bayesian probabilistic program, which represents probabilistic generative models expressed as structured procedures in an abstract description language \cite{lake2015human}. To achieve the desired performance, DI-Parser seamlessly integrates three key concepts: compositionality, transferability, and learning to learn. 

As a compositional program, DI-Parser is constructed from four basic components: intents, intent distributions, intent transitions, and neural embeddings. The Bayesian combinations of these components effectively capture the hidden structures of natural language dialogues, facilitating natural transfer across different domains. Furthermore, the ``learning to learn'' mechanism enables DI-Parser to incorporate prior experiences with related intents as hierarchical priors. By leveraging this framework, DI-Parser can identify a broad range of latent intents across various domains, even when each domain contains only a few labeled examples. This capability aligns with the principles of few-shot learning methods \cite{feng2024autodemopromptingleveraginggenerated}.

To sum up, the main contributions of this paper are as follows: 

\begin{itemize}
    \item We introduce the ``DI-Parser,'' a Neural-Bayesian Program Learning model for few-shot dialogue intent parsing that pioneers the decomposition of latent dialogue structures into several compositional components that are transferable across different scenarios.
    
    \item We propose a novel ``learning to learn'' framework, which, to the best of our knowledge, represents the first application of this approach to the intent parsing task utilizing the Neural-Bayesian Program Learning (NBPL) model.
    
    \item We evaluate the proposed method on a real-life industrial dataset, demonstrating a significant performance improvement over state-of-the-art methods. The DI-Parser achieves an accuracy that is 1.37\% higher than the Transformer model, marking one step closer to the human-reported inter-annotator agreement.
    
\end{itemize}

The rest of the paper is organized as follows. We first briefly review the related work in Section~\ref{sec:related}. Then, we discuss the details of the proposed DI-Parser in Section~\ref{sec:method}, followed by the experimental evaluations in Section~\ref{sec:exp}. Finally, we conclude this paper in Section~\ref{sec:con}.

\section{Related Work}
\label{sec:related}

\subsection{Bayesian Program Learning} 

Bayesian Program Learning (BPL) is a well-established machine learning approach that employs probabilistic models to express and learn algorithms \cite{10.1145/3308560.3316582}. By treating programs as probabilistic models and inferring their behavior from observed data, this method learns complex concepts from limited datasets while effectively capturing the generative processes underlying the data \cite{lake2015human,liang2010learning}. A key aspect of BPL is its ability to encode domain knowledge into the generative model, which enables effective learning from a limited number of training instances. 

For instance, the BPL was first applied to model the generative process of human handwritten characters, achieving human-level concept learning for character classification \cite{lake2015human}. To improve the model's ability to generalize and make predictions based on the learned programs, BPL typically employs model inference methods such as heuristic search-based approaches and Markov Chain Monte Carlo (MCMC)-based paradigms, which are essential for approximating the true posterior distributions over programs \cite{schkufza2013stochastic,ellis2016sampling}.

\subsection{Dialogue Intent Parsing}

Dialogue intent parsing involves extracting the user's intent from conversational utterances, which is crucial for building effective dialogue systems and achieving accurate natural language understanding (NLU). Traditional classification approaches include Naive Bayes \cite{grau2004dialogue}, Hidden Markov Models (HMM) \cite{stolcke2000dialogue}, and SVM-HMM \cite{tavafi2013dialogue}, which provide a foundational baseline for this field. Various deep learning techniques have advanced the accuracy of intent parsing by capturing intricate linguistic patterns, including models like Recurrent Neural Networks (RNNs), Long Short-Term Memory networks (LSTMs), and Transformer-based architectures \cite{deng2012use,yann2014zero}.

Compared to human perception, deep learning models require large amounts of training data and often suffer from poor sample efficiency. Few-shot learning aims to address these issues, with current approaches including matching networks \cite{vinyals2016matching}, prototypical networks \cite{snell2017prototypical}, and model-agnostic meta-learning \cite{finn2017model}. Additionally, few-shot Large Language Models (LLMs) have demonstrated remarkable effectiveness across various natural language processing tasks \cite{song2023communication, dial-in-llm, llm-topic-model}. These methods are designed to enable models to make accurate predictions with limited training data by leveraging meta-learning strategies that allow the model to learn how to learn from past experiences on different tasks \cite{song2020topicocean}.

Pre-trained models can also be effectively leveraged to facilitate high classification accuracy with limited training data \cite{infantcrynet}. Transfer learning plays a significant role in this domain, allowing models to benefit from knowledge gained from solving one problem and applying it to a different but related problem \cite{devlin-etal-2019-bert, wu2023enhance, song2019topic}. This approach is particularly useful in natural language understanding (NLU) tasks where large annotated datasets are scarce.


\section{Methodology}
\label{sec:method}

The key aspect of the DI-Parser is its use of ``learning to learn'' mechanism and unsupervised learning to estimate the dialogue act distribution relevant to the problem at hand, as well as the transition probabilities of these acts. A significant challenge for the analytic model is the dynamic discovery of new intents through unsupervised learning. In this section, we introduce the DI-Parser from two perspectives: the compositional components and the neural-Bayesian program framework.

\begin{figure*}[t!]
    \centering
    \includegraphics[width=0.7\textwidth]{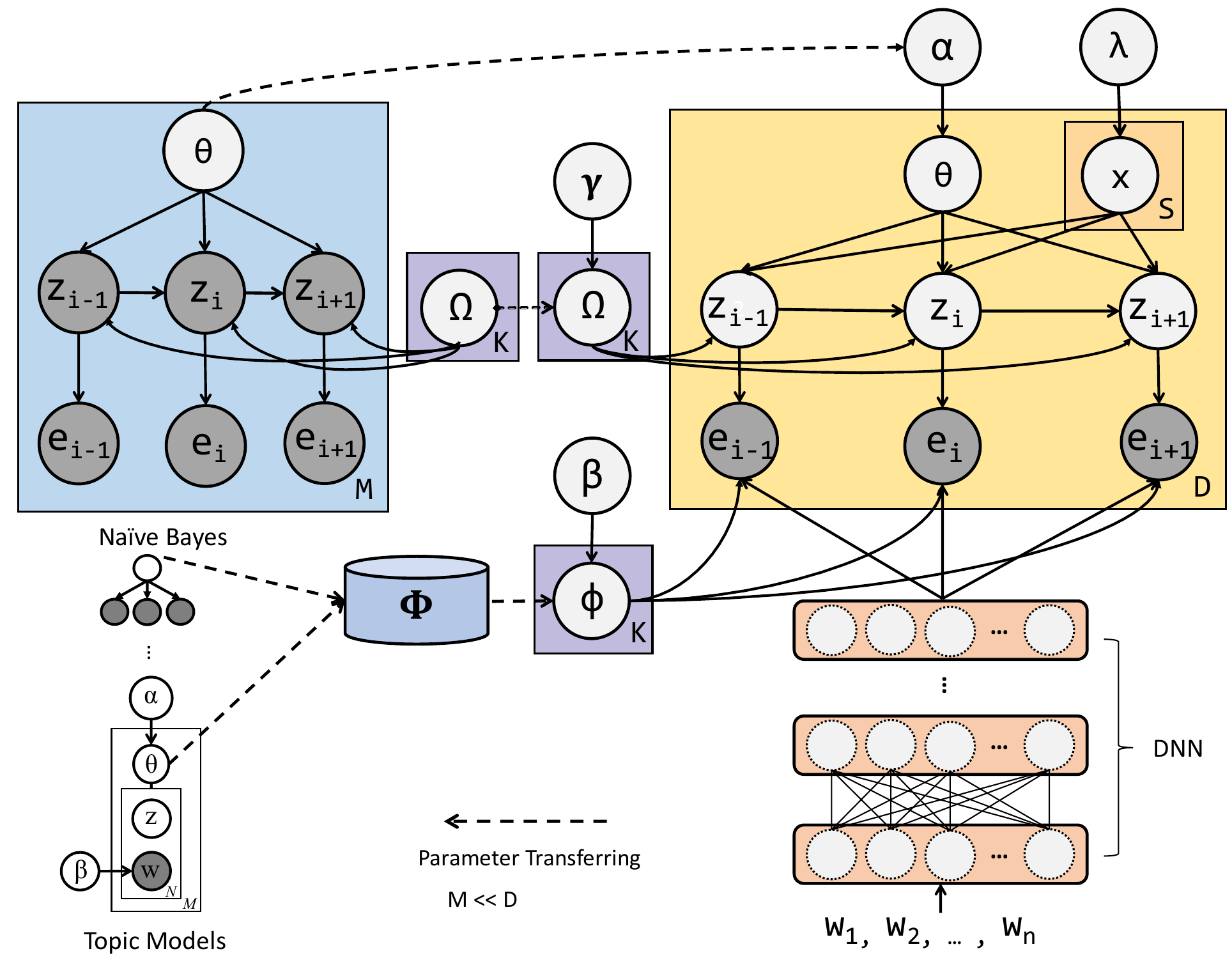}
    \caption{Graphical model of our proposed DI-Parser, whose learning process is initialized by learning to learn.}
    \label{fig:model}
\end{figure*}

\subsection{Compositional Components of DI-Parser}

In this section, we present the four basic components of DI-Parser: intent, intent distribution, intent transition matrix, and neural embedding.

\subsubsection{Intent}

The \textbf{intent} \( z_j \) is modeled as a Multinomial distribution over words, represented in the format \(\{(w_1, p_{w_1}^{z_j}), (w_2, p_{w_2}^{z_j}), \cdots\}\), where \( w_i \) denotes the word ID, \( z_j \) denotes the topic ID, and \( p_{w_i}^{z_j} \) represents the normalized weight of word \( w_i \) under topic \( z_j \). The Multinomial distribution can be derived from the \( P(w|c) \) distribution in Naive Bayes or the \( p(w|z) \) distribution in probabilistic topic models \cite{jiang2023probabilistic}, where \( c \) indicates the class tag and \( z \) is the latent topic. In this paper, we gather the intents from these two sources. For labeled data with intent tags, we train a Naive Bayes classifier and collect the \( P(w|c) \) distributions. For data lacking intent tag information, we train topic models and collect the \( p(w|z) \) distributions.

To maintain the compactness of the Intent Set, only intents that are sufficiently distinct from existing intents will be added to the Intent Set. Inspired by \cite{jiang2019federated}, we define the similarity measure between two topics \( z_i \) and \( z_j \) as


\begin{equation}
\label{eq:w_jac}
\resizebox{\columnwidth}{!}{$\begin{aligned}\rho(z_i,z_j) = & \frac{\sum_{t=1}^m \min(p_{w_t}^{z_i}, p_{w_t}^{z_j})}{\sum_{t=1}^m \max(p_{w_t}^{z_i}, p_{w_t}^{z_j}) + \sum_{t=m+1}^T p_{w_t}^{z_i} + \sum_{t=m+1}^T p_{w_t}^{z_j}} \\
= & \frac{\sum_{t=1}^m \min(p_{w_t}^{z_i}, p_{w_t}^{z_j})}{\sum_{t=1}^T p_{w_t}^{z_i} + \sum_{t=1}^T p_{w_t}^{z_j} - \sum_{t=1}^m \min(p_{w_t}^{z_i}, p_{w_t}^{z_j})}\end{aligned}$}
\raisetag{1\baselineskip}
\end{equation}

where \( P_{z_{\cdot}} \) denotes the top-\( T \) words distribution of topic \( z_{\cdot} \), with the detailed definition \( P_{z_{\cdot}} = (p_{w_1}^{z_{\cdot}}, p_{w_2}^{z_{\cdot}}, \cdots, p_{w_m}^{z_{\cdot}}, p_{w_{m+1}}^{z_{\cdot}}, \cdots, p_{w_T}^{z_{\cdot}}) \). The number of common words in their top-\( T \) words is represented by \( m \) (where \( 0 \leq m \leq T \)). Two topics are regarded as redundant if their similarity \( \rho(z_i, z_j) \) exceeds the predefined threshold \( \delta \).

\subsubsection{Intent Distribution}

The \textbf{intent distribution} captures the long-range dependency of intents within each dialogue and precisely reflects the characteristics of the domain from which the dialogues are generated. For example, in a market sales support service domain, the intent distribution mainly focuses on intents such as ``price inquiry'', while in a company management domain, it emphasizes intents such as ``room booking'' and ``recruitment''. The intent distribution, represented as \( \theta \) over all the intents, is modeled as a Multinomial discrete distribution with a Dirichlet process prior \( \alpha \).

\subsubsection{Intent Transition}

The \textbf{intent transition} captures the local dependency between intents within each dialogue and is modeled by a 2-dimensional matrix \( \mathcal{M} \), where each element \( m_{i,j} \) represents the probability of transitioning from intent \( z_i \) to intent \( z_j \).

\subsubsection{Neural Embedding}

The DI-Parser employs \textbf{neural embedding} \( \mathbf{e} \) to represent each utterance \( u \) using a network similar to the method described in \cite{le2014distributed}. The embedding distribution for each intent \( z \) is mathematically modeled by a Gaussian distribution \( \mathcal{N}(\mathbf{e}|\mu_{\mathbf{z}}, \sigma_{\mathbf{z}}^2) \). It is noteworthy that other methods, such as Word2Vec \cite{mikolov2013distributed}, can also be utilized in this context.

\subsection{Neural-Bayesian Program of DI-Parser}

The NBPL approach learns stochastic programs to represent concepts \cite{lake2015human}, constructing them compositionally from components of intents and utterances. The joint distribution over intents \( I_1, \ldots, I_n \) and the corresponding dialogues \( d_1, \ldots, d_n \) is expressed as follows:

\vspace{-1em}

\begin{equation}
\resizebox{\columnwidth}{!}{
$P(I_1, \cdots, I_n, d_1, \cdots, d_n)
= P(I_1) P(d_1|I_1) \prod_{i=2}^{m} P(d_i|I_i) P(I_i|I_{i-1})$}
\label{eq:bpl1}
\end{equation}

The generative process is described in  Algorithm \ref{alg:DI-Parser} with the graphical model illustrated in Figure \ref{fig:model}. The DI-Parser initializes the $\vec{\alpha}$, $\vec{\beta}$, $\tau$ and $\Omega$ using few-shot learning exemplars. For each intent $z_i$, the DI-Parser initializes the word distribution $\phi_{i}$.
For each multi-turn dialogue $d$ in the corpus $D$, the algorithm generates the intent distribution $\theta_d$ with a Dirichlet prior parameterized by $\vec{\alpha}$. For each utterance in this dialogue, the specific intent can either be drawn conditioned on previous intent $z_{i-1}$ from a Multinomial distribution using the Intent Transition matrix, or from the Multinomial intent distribution $\theta_{d}$.

\subsection{Parsing Dialogues with DI-Parser}

In this section, we describe how to parse the intent for each utterances given the dialogue transcript.

\subsubsection{Parsing without Learning to Learn}
\label{sec:training}

The objective of inference is to maximize the likelihood of the observed utterance $\mathbf{u}$: $P(\mathbf{u}|\alpha, \beta, \tau, \mu, \sigma, \Omega)$. Since evaluating the likelihood is intractable, it is infeasible to compute the optimal $z$ by maximizing $P(\mathbf{u}|\alpha, \beta, \tau, \mu, \sigma, \Omega)$ directly. Instead, we compute the complete likelihood $P(\mathbf{u}, \mathbf{i}, \mathbf{z}|\alpha, \beta, \tau, \mu, \sigma, \Omega)$:

\vspace{-1em}

\begin{equation}
\resizebox{\columnwidth}{!}{$
\begin{aligned}
P(\mathbf{u}, \mathbf{i}, \mathbf{z}|\alpha, \beta,\tau, \mu, \sigma, \Omega )
= P(\mathbf{z}|\alpha, \Omega, \mathbf{i} )P(\mathbf{i}|\tau)P(\mathbf{u} | \mathbf{z}, \mu, \sigma ) \\
= (1 - \tau)^{A} \tau^{B} \times \mathcal{N}(\mathbf{u}|\mu_{\mathbf{z}}, \sigma_{\mathbf{z}}^2) \\
\times (\frac{\Gamma (\sum_{z=1}^{T} \alpha_z)}{\Gamma (\prod_{z=1}^{T} \Gamma (\alpha_z))})^{D} \prod_{d=1}^{D}
(\frac{\prod_{z=1}^{T}\Gamma (C_{dz}+ \alpha_z)}{\sum_{z=1}^{T} (C_{dz}+ \alpha_z)}) \\
\times (\frac{\Gamma (\sum_{z=1}^{T} \alpha^{'}_z)}{\Gamma (\prod_{z=1}^{T} \Gamma (\alpha^{'}_z))})^{D} \\
\prod_{d=1}^{D} (\frac{\prod_{z=1}^{T}\Gamma (C_{dz}+ \alpha_z^{'})}{\sum_{z=1}^{T} (C_{dz}+ \alpha_z^{'})} (\frac{\prod_{z=1}^{T}\Gamma (C_{dz}+ \alpha_z^{'} + \delta (z_{i-1}=k=z_{i+1}) )}{\sum_{z=1}^{T} (C_{dz}+ \alpha_z^{'} + \delta (z_{i-1}=k))})
\end{aligned}$}
\end{equation}

\noindent Let \( C_{dz} \) denote the number of utterances assigned to intent \( z \) in dialogue \( d \). Additionally, let \( C_{z_{i-1},k}^{-i} \) represent the count of the \( i \)-th sentences assigned to topic \( k \), given that the \( (i-1) \)-th sentences are assigned to topic \( z_{i-1} \). The variables \( A \) and \( B \) correspond to the counts of \( 0 \) and \( 1 \) generated by the Bernoulli distribution, respectively. By applying Bayes' theorem, the full conditional assignment of intent \( k \) to \( z_{ds} \) can be expressed as follows:

\begin{equation}
\resizebox{\columnwidth}{!}{$
\begin{aligned}
P(z_{du} = k, \mathbf{i}_{du} | \mathbf{u}, \mathbf{z}_{du}, \alpha, \beta, \tau, \mu, \sigma, \Omega  )
= \mathcal{N}(\mathbf{u}|\mu_{\mathbf{z}}, \sigma_{\mathbf{z}}^2) \\
\prod_{u \in D\,  \& \, i_\mathbf{u} = 1}  (\frac{C_{z_{i-1},k}^{-i}+\alpha^{'}_{z}}{C_{z_{i-1, \cdot}}^{-i}+K\alpha^{'}_{z}}
\frac{C_{k,z_{i+1}}^{-i}+\alpha^{'}_{z}+\delta (z_{i-1}=k=z_{i+1})}{C_{k,\cdot }^{-i}+K\alpha^{'}_{z}+\delta (z_{i-1}=k)}) \\
\prod_{u \in D\,  \& \, i_\mathbf{u} = 0} \frac{C_{dz} + \alpha_k}{\sum_{k^{'}=1}^{K}(C_{dk^{'}} + \alpha_{k^{'}})} \times  (1 - \tau)^{A} \tau^{B}
\end{aligned}$}
\end{equation}


\noindent Similar to \cite{nguyen2015improving}, we approximate the above equation and integrate out \( \mathbf{i}_{ds} \):

\begin{equation}
\resizebox{\columnwidth}{!}{$
\begin{aligned}
P(z_{du} = k | \mathbf{u}, \mathbf{z}_{du}, \mathbf{i}_{du}, \alpha, \beta, \tau, \mu, \sigma, \Omega  ) \approx \mathcal{N}(\mathbf{u}|\mu_{\mathbf{z}}, \sigma_{\mathbf{z}}^2) \\
 ((1 - \tau) (\frac{C_{z_{i-1},k}^{-i}+\alpha^{'}_{z}}{C_{z_{i-1, \cdot}}^{-i}+K\alpha^{'}_{z}}
\frac{C_{k,z_{i+1}}^{-i}+\alpha^{'}_{z}+\delta (z_{i-1}=k=z_{i+1})}{C_{k,\cdot }^{-i}+K\alpha^{'}_{z}+\delta (z_{i-1}=k)}) \\
\tau \frac{C_{dz} + \alpha_k}{\sum_{k^{'}=1}^{K}(C_{dk^{'}} + \alpha_{k^{'}})})
\end{aligned}$}
\end{equation}

\begin{algorithm}[t!]
  \Begin{
    $\vec{\alpha}$, $\vec{\beta}$, $\tau$ and $\Omega$ are initialized with few-shot learning exemplar\\
    \For{each intent $z_i$}{
       draw a word distribution $\phi_i \sim Multinomial(\vec{\beta})$ \\
       draw a transition distribution $\Omega_i \sim Multinomial(\vec{\rho})$
    }
    \For{each dialogue $d \in D$}{
      draw intent distribution $\theta_d \sim Dirichlet(\vec{\alpha})$ \\
      \For {each utterance $u \in d$} {
        draw an indicator $x \sim Bernoulli(\tau)$ \\
        \If{$x=1$}{
        draw an intent $z_{ds} \sim \theta_{d}$ \\
        }\Else{
        draw an intent $z_{ds} \sim Multinomial(\Omega_{z_{ds}^{old}})$ \\
        }
      draw the utterance $\mathbf{e}_i \sim \mathcal{N}(\mathbf{u}|\mu_{\mathbf{z}}, \sigma_{\mathbf{z}}^2)$
     }
    }
  }
  \caption{DI-Parser}
  \label{alg:DI-Parser}
\end{algorithm}

\noindent For each sentence $s$, its latent indicator $x_{s}$ is sampled as follows:

\vspace{-1.2em}
\begin{equation}
\begin{aligned}
p(x=0|z)=(1-\tau) \ast \frac{C_{z_{i-1},z}^{-i}+\alpha^{'}_{z}}{C_{z_{i-1},\cdot }^{-i}+K\alpha^{'}_{z}}
\end{aligned}
\end{equation}

\begin{equation}
\begin{aligned}
p(x=1|z)= \tau \ast  \frac{C_{d,z}^{-i}+\alpha }{C_{d,\cdot }^{-i}+K\alpha}
\end{aligned}
\end{equation}

\subsubsection{Parsing with Learning to Learn}

We employ ``learning to learn'' mechanism to alleviate the burden of training a DI-Parser from three aspects: Intent, Intent Distribution, and Intent Transition Matrix. This approach benefits intent parsing by transferring the Intent Set as an initialization to enhance inference for new data. The Dirichlet prior \( \alpha \) of the Intent Distribution \( \theta \) and the Intent Transition Matrix \( \omega \) are both initialized using the distribution inferred from the few labeled instances. Following this transfer, the Gibbs sampling formula \( P(z_{s}=k | \mathbf{z_{-s}}, \mathbf{w}) \) is updated to:

\begin{equation}
\resizebox{\columnwidth}{!}{$
\begin{aligned}
P(z_{du} = k | \mathbf{u}, \mathbf{z}_{du}, \mathbf{i}_{du}, \alpha, \beta, \tau, \mu, \sigma, \Omega  ) \approx \mathcal{N}(\mathbf{u}|\mu_{\mathbf{z}}, \sigma_{\mathbf{z}}^2) \\
 ((1 - \tau) (\frac{C_{z_{i-1},k}^{-i}+\Omega^{LD}_{z}}{C_{z_{i-1, \cdot}}^{-i}+K\Omega^{LD}_{z}}
\frac{C_{k,z_{i+1}}^{-i}+\Omega^{LD}_{z}+\delta (z_{i-1}=k=z_{i+1})}{C_{k,\cdot }^{-i}+K\Omega^{LD}_{z}+\delta (z_{i-1}=k)}) \\
\tau \frac{C_{dz} + \theta^{LD}}{\sum_{k^{'}=1}^{K}(C_{dk^{'}} + \theta^{LD}_{k^{'}})} )
\end{aligned}$}
\end{equation}

\noindent where $\theta^{LD}$ and $\Omega^{LD}$ are priors learned from few labeled instances.

\section{Experiments}
\label{sec:exp}

In this section, we evaluate our DI-Parser on a real-life industrial dataset and demonstrate its advantages compared to other widely used methods.

\subsection{Dataset}
\label{sec:data_preparation}

The dataset was collected from real-world customer service interactions within the financial domain and was transcribed using an audio-to-text method \cite{song2021smartmeeting, song2021multimodal, song2022platform, jiang2021gdpr}. It contains a total of 952,749 dialogues. We utilized 80\% of the data for compiling the Intent Set, reserving the remaining 20\% for testing. By leveraging the ``wisdom of the crowd'' through the crowdsourcing platform Amazon Mechanical Turk (AMT) \cite{10184732, zhang2020cleaning}, a subset of 1,000 dialogues was labeled by human annotators, providing a rich foundation for few-shot learning.









\subsection{Experimental Results}

Table~\ref{tab:results} compares the results obtained using our DI-Parser model with those of previous methods. The results indicate that the proposed model ooffers promising accuracy, achieving an improvement of 1.7\% over the state-of-the-art Transformer model.


\begin{table}[t]
\centering
\begin{tabular}{r|c}
  \toprule
  \textbf{Model} & \textbf{Acc(\%)} \\
  \midrule
  Logistic Regression &  64.3  \\
  Hierarchical Attention Network & 72.6 \\
  Seq2Seq With Attention & 74.2 \\
  Transformer & 77.9 \\
  \textbf{DI-Parser} & \textbf{79.6} \\
  \bottomrule
\end{tabular}
\caption{Performance comparison in intent parsing}
\label{tab:results}
\end{table}

\section{Conclusion}
\label{sec:con}
In this paper, we introduced DI-Parser, a novel Neural-Bayesian Program Learning model designed for the implementation of dialogue intent parsing systems in a data-hungry setting. DI-Parser leverages data from multiple sources in a "learning to learn" framework, effectively harnessing the "wisdom of the crowd" through few-shot learning capabilities. Experimental results demonstrate that DI-Parser consistently outperforms traditional deep learning approaches in real-world scenarios, highlighting its practical advantages. Overall, this paper provides an effective solution to few-shot intent parsing and establishes a solid theoretical foundation for further research in few-shot text mining.

\bibliography{custom}

\begin{thebibliography}{46}
\providecommand{\natexlab}[1]{#1}

\bibitem[{Cui et~al.(2019)Cui, Lian, Jiang, Song, Bao, and Jiang}]{cui2019dal}
Shaobo Cui, Rongzhong Lian, Di~Jiang, Yuanfeng Song, Siqi Bao, and Yong Jiang. 2019.
\newblock Dal: Dual adversarial learning for dialogue generation.
\newblock \emph{arXiv preprint arXiv:1906.09556}.

\bibitem[{Deng et~al.(2012)Deng, Tur, He, and Hakkani-Tur}]{deng2012use}
Li~Deng, Gokhan Tur, Xiaodong He, and Dilek Hakkani-Tur. 2012.
\newblock Use of kernel deep convex networks and end-to-end learning for spoken language understanding.
\newblock In \emph{SLT}.

\bibitem[{Devlin et~al.(2019)Devlin, Chang, Lee, and Toutanova}]{devlin-etal-2019-bert}
Jacob Devlin, Ming-Wei Chang, Kenton Lee, and Kristina Toutanova. 2019.
\newblock \href {https://doi.org/10.18653/v1/N19-1423} {{BERT}: Pre-training of deep bidirectional transformers for language understanding}.
\newblock In \emph{Proceedings of the 2019 Conference of the North {A}merican Chapter of the Association for Computational Linguistics: Human Language Technologies, Volume 1 (Long and Short Papers)}, pages 4171--4186, Minneapolis, Minnesota. Association for Computational Linguistics.

\bibitem[{Ellis et~al.(2016)Ellis, Solar-Lezama, and Tenenbaum}]{ellis2016sampling}
Kevin Ellis, Armando Solar-Lezama, and Josh Tenenbaum. 2016.
\newblock Sampling for bayesian program learning.
\newblock In \emph{NIPS}.

\bibitem[{Feng et~al.(2024)Feng, Hong, and Zhang}]{feng2024autodemopromptingleveraginggenerated}
Longyu Feng, Mengze Hong, and Chen~Jason Zhang. 2024.
\newblock \href {https://arxiv.org/abs/2410.01724} {Auto-demo prompting: Leveraging generated outputs as demonstrations for enhanced batch prompting}.
\newblock \emph{Preprint}, arXiv:2410.01724.

\bibitem[{Finn et~al.(2017)Finn, Abbeel, and Levine}]{finn2017model}
Chelsea Finn, Pieter Abbeel, and Sergey Levine. 2017.
\newblock Model-agnostic meta-learning for fast adaptation of deep networks.
\newblock In \emph{ICML}.

\bibitem[{Firdaus et~al.(2018)Firdaus, Bhatnagar, Ekbal, and Bhattacharyya}]{firdaus2018deep}
Mauajama Firdaus, Shobhit Bhatnagar, Asif Ekbal, and Pushpak Bhattacharyya. 2018.
\newblock A deep learning based multi-task ensemble model for intent detection and slot filling in spoken language understanding.
\newblock In \emph{ICONIP}.

\bibitem[{Grau et~al.(2004)Grau, Sanchis, Castro, and Vilar}]{grau2004dialogue}
Sergio Grau, Emilio Sanchis, Maria~Jose Castro, and David Vilar. 2004.
\newblock Dialogue act classification using a bayesian approach.
\newblock In \emph{9th Conference Speech and Computer}.

\bibitem[{Harrigan et~al.(2018)Harrigan, Evers, Miles, and Daly}]{harrigan2018customer}
Paul Harrigan, Uwana Evers, Morgan~P Miles, and Tim Daly. 2018.
\newblock Customer engagement and the relationship between involvement, engagement, self-brand connection and brand usage intent.
\newblock \emph{Journal of Business Research}, 88:388--396.

\bibitem[{Ji et~al.(2016)Ji, Haffari, and Eisenstein}]{ji2016latent}
Yangfeng Ji, Gholamreza Haffari, and Jacob Eisenstein. 2016.
\newblock A latent variable recurrent neural network for discourse relation language models.
\newblock \emph{arXiv preprint arXiv:1603.01913}.

\bibitem[{Jiang et~al.(2016)Jiang, Shi, Lian, and Wu}]{jiang2016latent}
Di~Jiang, Lei Shi, Rongzhong Lian, and Hua Wu. 2016.
\newblock Latent topic embedding.
\newblock In \emph{Proceedings of COLING 2016, the 26th International Conference on Computational Linguistics: Technical Papers}, pages 2689--2698.

\bibitem[{Jiang et~al.(2021{\natexlab{a}})Jiang, Song, Lian, Bao, Peng, He, Wu, Zhang, and Chen}]{jiang2021familia}
Di~Jiang, Yuanfeng Song, Rongzhong Lian, Siqi Bao, Jinhua Peng, Huang He, Hua Wu, Chen Zhang, and Lei Chen. 2021{\natexlab{a}}.
\newblock Familia: A configurable topic modeling framework for industrial text engineering.
\newblock In \emph{Database Systems for Advanced Applications: 26th International Conference, DASFAA 2021, Taipei, Taiwan, April 11--14, 2021, Proceedings, Part III 26}, pages 516--528. Springer.

\bibitem[{Jiang et~al.(2019)Jiang, Song, Tong, Wu, Zhao, Xu, and Yang}]{jiang2019federated}
Di~Jiang, Yuanfeng Song, Yongxin Tong, Xueyang Wu, Weiwei Zhao, Qian Xu, and Qiang Yang. 2019.
\newblock Federated topic modeling.
\newblock In \emph{CIKM}.

\bibitem[{Jiang et~al.(2021{\natexlab{b}})Jiang, Tan, Peng, Chen, Wu, Zhao, Song, Tong, Liu, Xu et~al.}]{jiang2021gdpr}
Di~Jiang, Conghui Tan, Jinhua Peng, Chaotao Chen, Xueyang Wu, Weiwei Zhao, Yuanfeng Song, Yongxin Tong, Chang Liu, Qian Xu, et~al. 2021{\natexlab{b}}.
\newblock A gdpr-compliant ecosystem for speech recognition with transfer, federated, and evolutionary learning.
\newblock \emph{ACM Transactions on Intelligent Systems and Technology (TIST)}, 12(3):1--19.

\bibitem[{Jiang et~al.(2015)Jiang, Vosecky, Leung, Yang, and Ng}]{jiang2015sg}
Di~Jiang, Jan Vosecky, Kenneth Wai-Ting Leung, Lingxiao Yang, and Wilfred Ng. 2015.
\newblock Sg-wstd: A framework for scalable geographic web search topic discovery.
\newblock \emph{Knowledge-Based Systems}, 84:18--33.

\bibitem[{Jiang and Yang(2016)}]{jiang2016query}
Di~Jiang and Lingxiao Yang. 2016.
\newblock Query intent inference via search engine log.
\newblock \emph{Knowledge and information systems}, 49:661--685.

\bibitem[{Jiang et~al.(2023)Jiang, Zhang, and Song}]{jiang2023probabilistic}
Di~Jiang, Chen Zhang, and Yuanfeng Song. 2023.
\newblock \emph{Probabilistic topic models: Foundation and application}.
\newblock Springer.

\bibitem[{Kato et~al.(2017)Kato, Nagai, Noda, Sumitomo, Wu, and Yamamoto}]{kato2017utterance}
Tsuneo Kato, Atsushi Nagai, Naoki Noda, Ryosuke Sumitomo, Jianming Wu, and Seiichi Yamamoto. 2017.
\newblock Utterance intent classification of a spoken dialogue system with efficiently untied recursive autoencoders.
\newblock In \emph{SIGdial}.

\bibitem[{Khanpour et~al.(2016)Khanpour, Guntakandla, and Nielsen}]{khanpour2016dialogue}
Hamed Khanpour, Nishitha Guntakandla, and Rodney Nielsen. 2016.
\newblock Dialogue act classification in domain-independent conversations using a deep recurrent neural network.
\newblock In \emph{COLING}.

\bibitem[{Lake et~al.(2015)Lake, Salakhutdinov, and Tenenbaum}]{lake2015human}
Brenden~M Lake, Ruslan Salakhutdinov, and Joshua~B Tenenbaum. 2015.
\newblock Human-level concept learning through probabilistic program induction.
\newblock \emph{Science}, 350(6266):1332--1338.

\bibitem[{Le and Mikolov(2014)}]{le2014distributed}
Quoc Le and Tomas Mikolov. 2014.
\newblock Distributed representations of sentences and documents.
\newblock In \emph{ICML}.

\bibitem[{Li et~al.(2021)Li, Jiang, Lian, Wu, Tan, Xu, and Su}]{li2021heterogeneous}
Yawen Li, Di~Jiang, Rongzhong Lian, Xueyang Wu, Conghui Tan, Yi~Xu, and Zhiyang Su. 2021.
\newblock Heterogeneous latent topic discovery for semantic text mining.
\newblock \emph{IEEE Transactions on Knowledge and Data Engineering}, 35(1):533--544.

\bibitem[{Liang et~al.(2010)Liang, Jordan, and Klein}]{liang2010learning}
Percy Liang, Michael~I Jordan, and Dan Klein. 2010.
\newblock Learning programs: A hierarchical bayesian approach.
\newblock In \emph{ICML}.

\bibitem[{Mengze et~al.(2024{\natexlab{a}})Mengze, Chen, Lingxiao, Yuanfeng, and Di}]{infantcrynet}
Hong Mengze, Jason~Zhang Chen, Yang Lingxiao, Song Yuanfeng, and Jiang Di. 2024{\natexlab{a}}.
\newblock {InfantCryNet}: A data-driven framework for intelligent analysis of infant cries.
\newblock In \emph{Proceedings of the 16th Asian Conference on Machine Learning}, volume 260 of \emph{Proceedings of Machine Learning Research}. PMLR.

\bibitem[{Mengze et~al.(2024{\natexlab{b}})Mengze, Di, Hanlin, Yuanfeng, Haijun, Lixin, and Chen}]{dial-in-llm}
Hong Mengze, Jiang Di, Gu~Hanlin, Song Yuanfeng, Yang Haijun, Fan Lixin, and Jason~Zhang Chen. 2024{\natexlab{b}}.
\newblock Dial-in llm: Leveraging few-shot large language models for human-aligned dialogue intent discovery and text clustering.
\newblock \emph{arXiv preprint}.

\bibitem[{Mengze et~al.(2024{\natexlab{c}})Mengze, Di, Yuanfeng, and Chen}]{llm-topic-model}
Hong Mengze, Jiang Di, Song Yuanfeng, and Jason~Zhang Chen. 2024{\natexlab{c}}.
\newblock Augmenting topic models with few-shot large language models.
\newblock \emph{arXiv preprint}.

\bibitem[{Mikolov et~al.(2013)Mikolov, Sutskever, Chen, Corrado, and Dean}]{mikolov2013distributed}
Tomas Mikolov, Ilya Sutskever, Kai Chen, Greg~S Corrado, and Jeff Dean. 2013.
\newblock Distributed representations of words and phrases and their compositionality.
\newblock In \emph{NIPS}.

\bibitem[{Nguyen et~al.(2015)Nguyen, Billingsley, Du, and Johnson}]{nguyen2015improving}
Dat~Quoc Nguyen, Richard Billingsley, Lan Du, and Mark Johnson. 2015.
\newblock Improving topic models with latent feature word representations.
\newblock \emph{Transactions of the Association for Computational Linguistics}, 3:299--313.

\bibitem[{Peng et~al.(2019)Peng, Ma, Jiang, and Wu}]{10.1145/3308560.3316582}
Jinhua Peng, Zongyang Ma, Di~Jiang, and Hua Wu. 2019.
\newblock \href {https://doi.org/10.1145/3308560.3316582} {Integrating bayesian and neural networks for discourse coherence}.
\newblock In \emph{Companion Proceedings of The 2019 World Wide Web Conference}, WWW '19, page 294–300, New York, NY, USA. Association for Computing Machinery.

\bibitem[{Perevalov et~al.(2019)Perevalov, Kurushin, Faizrakhmanov, and Khabibrakhmanova}]{perevalov2019question}
Aleksandr Perevalov, Daniil Kurushin, Rustam Faizrakhmanov, and Farida Khabibrakhmanova. 2019.
\newblock Question embeddings based on shannon entropy: Solving intent classification task in goal-oriented dialogue system.
\newblock \emph{arXiv preprint arXiv:1904.00785}.

\bibitem[{Schkufza et~al.(2013)Schkufza, Sharma, and Aiken}]{schkufza2013stochastic}
Eric Schkufza, Rahul Sharma, and Alex Aiken. 2013.
\newblock Stochastic superoptimization.
\newblock In \emph{ACM SIGPLAN Notices}, volume~48, pages 305--316. ACM.

\bibitem[{Snell et~al.(2017)Snell, Swersky, and Zemel}]{snell2017prototypical}
Jake Snell, Kevin Swersky, and Richard Zemel. 2017.
\newblock Prototypical networks for few-shot learning.
\newblock In \emph{NIPS}.

\bibitem[{Song(2019)}]{song2019topic}
Yuanfeng Song. 2019.
\newblock Topic-aware dialogue speech recognition with transfer learning.
\newblock In \emph{Proceedings of the Annual Conference of the International Speech Communication Association, INTERSPEECH}.

\bibitem[{Song et~al.(2023)Song, He, Zhao, Gu, Jiang, Yang, Fan, and Yang}]{song2023communication}
Yuanfeng Song, Yuanqin He, Xuefang Zhao, Hanlin Gu, Di~Jiang, Haijun Yang, Lixin Fan, and Qiang Yang. 2023.
\newblock A communication theory perspective on prompting engineering methods for large language models.
\newblock \emph{arXiv preprint arXiv:2310.18358}.

\bibitem[{Song et~al.(2021{\natexlab{a}})Song, Huang, Zhao, Jiang, and Wong}]{song2021multimodal}
Yuanfeng Song, Xiaoling Huang, Xuefang Zhao, Di~Jiang, and Raymond Chi-Wing Wong. 2021{\natexlab{a}}.
\newblock Multimodal n-best list rescoring with weakly supervised pre-training in hybrid speech recognition.
\newblock In \emph{2021 IEEE International Conference on Data Mining (ICDM)}, pages 1336--1341. IEEE.

\bibitem[{Song et~al.(2021{\natexlab{b}})Song, Jiang, Zhao, Huang, Xu, Wong, and Yang}]{song2021smartmeeting}
Yuanfeng Song, Di~Jiang, Xuefang Zhao, Xiaoling Huang, Qian Xu, Raymond Chi-Wing Wong, and Qiang Yang. 2021{\natexlab{b}}.
\newblock Smartmeeting: Automatic meeting transcription and summarization for in-person conversations.
\newblock In \emph{Proceedings of the 29th ACM International Conference on Multimedia}, pages 2777--2779.

\bibitem[{Song et~al.(2022)Song, Lian, Chen, Jiang, Zhao, Tan, Xu, and Wong}]{song2022platform}
Yuanfeng Song, Rongzhong Lian, Yixin Chen, Di~Jiang, Xuefang Zhao, Conghui Tan, Qian Xu, and Raymond Chi-Wing Wong. 2022.
\newblock A platform for deploying the tfe ecosystem of automatic speech recognition.
\newblock In \emph{Proceedings of the 30th ACM International Conference on Multimedia}, pages 6952--6954.

\bibitem[{Song et~al.(2020)Song, Tong, Bao, Jiang, Wu, and Wong}]{song2020topicocean}
Yuanfeng Song, Yongxin Tong, Siqi Bao, Di~Jiang, Hua Wu, and Raymond Chi-Wing Wong. 2020.
\newblock Topicocean: An ever-increasing topic model with meta-learning.
\newblock In \emph{2020 IEEE International Conference on Data Mining (ICDM)}, pages 1262--1267. IEEE.

\bibitem[{Stolcke et~al.(2000)Stolcke, Ries, Coccaro, Shriberg, Bates, Jurafsky, Taylor, Martin, Ess-Dykema, and Meteer}]{stolcke2000dialogue}
Andreas Stolcke, Klaus Ries, Noah Coccaro, Elizabeth Shriberg, Rebecca Bates, Daniel Jurafsky, Paul Taylor, Rachel Martin, Carol~Van Ess-Dykema, and Marie Meteer. 2000.
\newblock Dialogue act modeling for automatic tagging and recognition of conversational speech.
\newblock \emph{Computational linguistics}.

\bibitem[{Tavafi et~al.(2013)Tavafi, Mehdad, Joty, Carenini, and Ng}]{tavafi2013dialogue}
Maryam Tavafi, Yashar Mehdad, Shafiq Joty, Giuseppe Carenini, and Raymond Ng. 2013.
\newblock Dialogue act recognition in synchronous and asynchronous conversations.
\newblock In \emph{SIGDIAL}.

\bibitem[{Vinyals et~al.(2016)Vinyals, Blundell, Lillicrap, Wierstra et~al.}]{vinyals2016matching}
Oriol Vinyals, Charles Blundell, Timothy Lillicrap, Daan Wierstra, et~al. 2016.
\newblock Matching networks for one shot learning.
\newblock In \emph{NIPS}.

\bibitem[{Wu et~al.(2023)Wu, Jiang, Song, Xu, and Yang}]{wu2023enhance}
Xueyang Wu, Di~Jiang, Yuanfeng Song, Qian Xu, and Qiang Yang. 2023.
\newblock Enhance mono-modal sentiment classification with federated cross-modal transfer.
\newblock \emph{IEEE Data Eng. Bull.}, 46(1):158--169.

\bibitem[{Yan et~al.(2017)Yan, Duan, Chen, Zhou, Zhou, and Li}]{yan2017building}
Zhao Yan, Nan Duan, Peng Chen, Ming Zhou, Jianshe Zhou, and Zhoujun Li. 2017.
\newblock Building task-oriented dialogue systems for online shopping.
\newblock In \emph{AAAI}.

\bibitem[{Yann et~al.(2014)Yann, Tur, Hakkani-Tur, and Heck}]{yann2014zero}
D~Yann, G~Tur, D~Hakkani-Tur, and L~Heck. 2014.
\newblock Zero-shot learning and clustering for semantic utterance classification using deep learning.

\bibitem[{Zhang et~al.(2020)Zhang, Zhang, Xie, Liu, Li, Jiang, Lin, Wu, and Chen}]{zhang2020cleaning}
Chen Zhang, Haodi Zhang, Weiteng Xie, Nan Liu, Qifan Li, Di~Jiang, Peiguang Lin, Kaishun Wu, and Lei Chen. 2020.
\newblock Cleaning uncertain data with crowdsourcing-a general model with diverse accuracy rates.
\newblock \emph{IEEE Transactions on Knowledge and Data Engineering}, 34(8):3629--3642.

\bibitem[{Zhang et~al.(2023)Zhang, Huang, Su, Chen, Jiang, Fan, Zhang, Lian, and Wu}]{10184732}
Haodi Zhang, Wenxi Huang, Zhenhan Su, Junyang Chen, Di~Jiang, Lixin Fan, Chen Zhang, Defu Lian, and Kaishun Wu. 2023.
\newblock \href {https://doi.org/10.1109/ICDE55515.2023.00099} {Hierarchical crowdsourcing for data labeling with heterogeneous crowd}.
\newblock In \emph{2023 IEEE 39th International Conference on Data Engineering (ICDE)}, pages 1234--1246.

\end{thebibliography}

\end{document}